\documentclass{article}
\usepackage[final]{bdl_2018}
\usepackage[utf8]{inputenc} 
\usepackage[T1]{fontenc}    
\usepackage{hyperref}       
\usepackage{url}            
\usepackage{booktabs}       
\usepackage{amsfonts}       
\usepackage{nicefrac}       
\usepackage{microtype}      
\usepackage{wrapfig}
\usepackage{graphicx}
\usepackage{color, colortbl}
\usepackage{comment}
\usepackage{natbib}

\title{Improving robustness of classifiers \\by training against live traffic}
\author{
  Kumar Sricharan, Kumar Saketh, Ashok Srivastava \\
  Central Data Science Organization, 
  Intuit Inc. \\
  \texttt{\{sricharan{\textunderscore}kumar, kumar{\textunderscore}kallurupalli, ashok{\textunderscore}srivastava\}@intuit.com}}

\begin{document}

\maketitle

\begin{abstract}
  Deep learning models are known to be overconfident in their predictions on out of distribution inputs. This is a challenge when a model is trained on a particular input dataset, but receives out of sample data when deployed in practice. Recently, there has been work on building building classifiers that are robust to out of distribution samples by adding a regularization term that maximizes the entropy of the classifier output on out of distribution data. However, given the challenge that it is not always possible to obtain out of distribution samples, the authors suggest a GAN based alternative that is independent of specific knowledge of out of distribution samples. From this existing work, we also know that having access to the true out of sample distribution for regularization works significantly better than using samples from the GAN. In this paper, we make the following observation: in practice, the out of distribution samples are contained in the traffic that hits a deployed classifier. However, the traffic will also contain a unknown proportion of in-distribution samples. If the entropy over of all of the traffic data were to be naively maximized, this will hurt the classifier performance on in-distribution data. To effectively leverage this traffic data, we propose an adaptive regularization technique (based on the maximum predictive probability score of a sample) which penalizes out of distribution samples more heavily than in distribution samples in the incoming traffic. This ensures that the overall performance of the classifier does not degrade on in-distribution data, while detection of out-of-distribution samples is significantly improved by leveraging the unlabeled traffic data. We show the effectiveness of our method via experiments on natural image datasets.
\end{abstract}

\section{Introduction}
Deep learning approaches have been shown to be susceptible to being overly confident on unseen instances that are outside of the training distribution. Recently, there have been number of techniques for accurate out of sample detection. This body of work can be divided into two broad categories: (i) methods that analyze the output of pre-trained classifier networks to detect out of sample distributions and (ii) methods that build more robust neural networks by modifying the loss function to include regularization terms such as priors over the weights or the output of the network. The work on using the maximum probability of the classifier~\cite{hendrycks2016baseline}, its modification to use temperature based scaling~\cite{liang2017principled}, and the work on using the underlying feature representations in combination with the Mahalanobis distance~\cite{lee2018simple} belong to category (i). Methods under category (ii) include Bayesian neural networks based on placing prior over weights of the network~\cite{hernandez2015probabilistic}, using dropout as a Bayesian approximation~\cite{li2017dropout}, ensemble based approaches~\cite{lakshminarayanan2017simple} and adding a regularization term to explicitly maximize entropy of out of distribution samples~\cite{lee2017training}. 

Our focus in this work is on the second category. In particular, we seek to build more robust classifiers that don't overfit on out of distribution samples by regularizing the classifier to explicitly lower the prediction confidence on out-of-distribution samples through the use of an entropy based regularization term. It has been shown~\cite{lee2017training} that when there is access to the representative out-of-distribution samples for regularization during training, this technique is extremely accurate (>99\%) \footnote{The accuracy is with respect to detection of out of distribution samples on a test set which has a even mixture with 50\% of in-distribution samples and 50\% of out of distribution samples.} in out-of-distribution detection. The challenge of course, is that it is not always possible to obtain the true out-of-distribution samples a classifier will encounter in the wild. This has prompted the development of a method to approximate the out of distribution samples in \cite{lee2017training}.  In particular,  this work uses a GAN to find samples that are close to training distribution that also have high entropy. The key intuition is that by maximizing the uncertainty of the classifier over these out of distribution samples close to the training distribution, the same effect will be propagated to all samples outside of the training distribution. This GAN based classifier, while significantly more accurate at detecting out-of-distribution samples than a standard classifier, falls well short of the >99\% accuracy that can be achieved with access to out-of-distribution samples.

\textbf{Contribution} Our first contribution is recognizing that while it is not possible to get the true out-of-distribution samples, we can approximate this by looking at the live traffic the classifier is encountering in practice. Specifically, we can collect samples of the traffic at suitably spaced periodic intervals, and periodically update the model by using this unlabeled traffic as a proxy for the out of distribution data the classifier will subsequently encounter before the next update. The challenge with this approach is that the traffic distribution will contain a mix of in-distribution and out-of-distribution samples, and naive regularization over this traffic sample could affect performance by virtue of forcing the classifier response to be uniformly spread across all classes on in-distribution data. 

To address this, our second contribution is to use a adaptive regularization scheme that selectively penalizes the out of distribution samples in the traffic while minimally affecting the classifier output on in-distribution samples. The adaptive regularization is based on using a crude detection score indicating probability of a sample being in-distribution vs out-of-distribution. The key to our method's effectiveness is that we need the detection score to only be roughly accurate and not perfect - this is because a small amount of regularization maximizing the entropy over the in-distribution has been shown to not affect, and in fact, improve classification performance~\cite{pereyra2017regularizing}. Our detection score is based on a squashed version of the max probability score, which has been shown to have useful signal for detecting out of distribution samples~\cite{hendrycks2016baseline}. The degree of squashing is controlled by a parameter that is tuned to ensure that performance on in-class data is not reduced, and if anything, is improved~\cite{pereyra2017regularizing}. We propose this adaptive regularization scheme next, and illustrate its effectiveness through experiments.

\section{Training robust classifiers using traffic data}
We start with the confidence loss proposed by Lee \emph{et.al.}\cite{lee2017training}:
    $L_c(\theta) = \mathbb{E}_{P_{in}(\hat{x},\hat{y})} [-\log \mathbb{P}_\theta (y = \hat{y} \mid \hat{x})] + \beta \mathbb{E}_{{P}_{out}(x)}[KL(\mathbb{U}(y) \mid\mid \mathbb{P}_\theta (y \mid {x}))]$
where the first term is the standard cross-entropy loss, and the second term forces the network (parameterized by $\theta$) to generate close to uniform distributions for out of distribution samples (represented by ${P}_{out}$). We note that the proposed loss function corresponds to a Bayesian classifier under a prior on the weights that corresponds to the classifier producing a uniform distribution on the network output of class probabilities $P_\theta(y \mid x)$.
\subsection{Adaptive regularization of traffic data}
The confidence loss function is very intuitive, but the key difficulty is in determining ${P}_{out}$. To circumvent this, we propose using the traffic data that a classifier receives in practice. Assume that the live traffic samples hitting a classifier are characterized by the distribution $P_{tf}$, which can be approximated by an unlabeled sample of the traffic data $D_{tf}$ that is collected for the purposes of training the classifier. If the traffic is not stationary, we note that the data set $D_{tf}$ and subsequently the classifier can be periodically updated such that the stationarity assumption holds between updates. 

To leverage traffic data, we modify the original confidence loss by first  replacing the out of distribution data $P_{out}$ with the sample traffic data $D_{tf}$. Next, To handle the fact that the samples in $D_{tf}$ contain a mix of in-distribution and out of distribution data, we introduce an adaptive regularization term $\beta(x)$ which is a function of $x$. The modified loss function is given by:
\begin{equation}
    L_{tf}(\theta) = \mathbb{E}_{P_{in}(\hat{x},\hat{y})} [-\log \mathbb{P}_\theta (y = \hat{y} \mid \hat{x})] +   \mathbb{E}_{{D}_{tf}(x)}[\beta(x) \times KL(\mathbb{U}(y) \mid\mid \mathbb{P}_\theta (y \mid {x}))]
\end{equation}
 This $\beta(x)$ term ideally should be $1$ for out of distribution samples in $D_{tf}$, and $0$ for in-distribution. However, this is a circular requirement - the very goal of this paper is to obtain such a detector!

\begin{wrapfigure}{R}{0.55\textwidth}
\vspace{-0.0in}
\includegraphics[width=.5\textwidth]{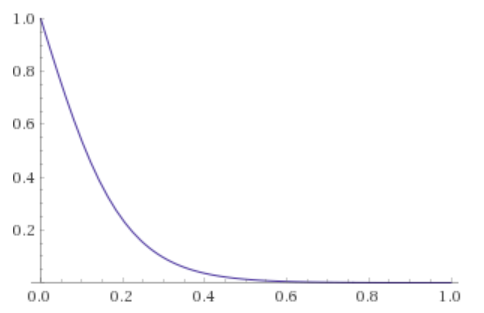}
\caption{\small{$\phi(z)$ for $\gamma=10$}}
\end{wrapfigure}

\subsection{Regularization term}
Instead, we propose to use an approximation to the ideal $\beta(x)$ that has some signal for distinguishing the in-distribution vs out-of-distribution samples. Specifically, we will use a non-linear transformation of the maximum probability score $m(x) = \max_{y} P(y|x)$ of the output of the classifier. The score $m(x)$ has been shown to have reasonable performance for detecting out of distribution samples in \cite{hendrycks2016baseline}. The specific non-linear transformation we propose to use is given by $\beta(x) = \phi(m(x))$, where $\phi(z) = 1 - 2({sigmoid}(\gamma(z)-0.5)$ is a squashing function. When training, during each iteration, we treat $\beta(x)$ as a fixed constant when computing the gradient of $L_{tf}(x)$\footnote{$\beta(x)$ is re-computed during each forward pass, but is treated as a constant during the backward pass}.

This squashing transformation allows for the regularization $\beta(x)$ to  penalize only $x$ for which the standard classifier is not confident. The degree of confidence is controlled by the parameter $\gamma$. The value for $\gamma$ is chosen in practice by using a small hold-out in-distribution dataset, and selecting the lowest value of $\gamma$ that does not result in a drop in performance over the hold out set~\cite{pereyra2017regularizing}.  Finally, we note that other alternative, approximate detectors can be used in place of $m(x)$. Next, we will show that this crude approximation to an ideal $\beta(x)$ performs remarkably well in practice. 


\section{Experimental results}
We showcase our results through experiments on CIFAR~\cite{krizhevsky2009learning} and SVHN~\cite{netzer2011reading}. We train VGGNet~\cite{szegedy2015going} for classification. Specifically, we take one dataset from \{CIFAR, SVHN\} as the in distribution dataset, and match with one dataset from \{CIFAR, SVHN\} as the out of sample dataset. For each (x-train, y-train) combination, we test using \{x-test\} to measure in-distribution classification performance, and (x-test, y-test) to measure detection accuracy. 
\subsection{Extreme scenarios} We start by considering two extremes for the traffic data: in the first scenario, we look at the case where the traffic data is exclusively the in-distribution data. In this case, to evaluate performance, we measure the the micro-averaged\footnote{\url{http://scikit-learn.org/stable/auto_examples/model_selection/plot_roc.html}} area under the precision and recall curve (AUPR)\footnote{We use AUPR over AUROC to account for the class-imbalance in the micro-average setting~\cite{saito2015precision}.} for multi-class classification over the test set of the in-class distribution.  In the second case, we look at the situation where the traffic data is exclusively the out-of-distribution data. For measuring out-of-distribution detection accuracy, we use baseline threshold-based detectors~\cite{hendrycks2016baseline} that computes the maximum value of predictive distribution on a test sample and classifies it as positive (i.e., in-distribution) if the confidence score is above some threshold. We compare the following set of classifiers: (i) $\beta=0$: no regularization, (ii) $\beta=1$\footnote{We tried different non-zero $\beta$ values (0.1, 0.2, 0.5), and it did not result in significant changes to results.}: fixed regularization, (iii) $\beta=\beta(x)$: adaptive regularization with $\gamma=10$, and (iv) GAN: GAN based regularization~\cite{lee2017training}.


\begin{table}[h]
\centering
\begin{small}
\scalebox{1.0}{
\begin{tabular}{c|cccc|cccc}
Dataset & \multicolumn{4}{c} {AUPR} & \multicolumn{4}{c} {Detection accuracy} \\
&$\beta=0$&$\beta=1$&$\beta=\beta(x)$&GAN&$\beta=0$&$\beta=1$&$\beta=\beta(x)$&GAN\\\hline
CIFAR-in/CIFAR-out &.877&{.792}&{.873}&.873 & 0.5&0.5&0.5 &0.5\\
SVHN-in/SVHN-out &.976&{.954}&{.975}&.974 & 0.5&0.5&0.5 &0.5\\\hline
CIFAR-in/SVHN-out &.877 &.872 &.874 &.873 & .680&.999&.999&.750 \\
SVHN-in/CIFAR-out &.976&.968 &.969 &.974 & .780 &.998 &.997 &.950 \\
\end{tabular}}
\end{small}
\vspace{0.2in}
\caption{The proposed adaptive regularization scheme performs close to the standard classifier wrt classification performance on in-distribution data, and close to an fixed regularization scheme trained on traffic data comprised exclusively of out-of-distribution data wrt out-of-distribution sample detection.}
\end{table}


From the first two rows, we note that when the traffic purely contains in-distribution data, the adaptive regularized classifier has performance close to a standard classifier with no regularization, unlike the fixed regularization classifier $\beta=1$. Additionally, from the last two rows where the traffic only has out-of-distribution data, we can see that it has near perfect detection accuracy (0.99+) like an oracle regularized classifier ($\beta=1$), and outperforms the GAN based method. 

\subsection{Even mixture} Next, we consider a more realistic situation where the traffic data has an even mixture of in-distribution and out-of-distribution samples. We repeat the experiment on this even mixture traffic data. We compare the following set of classifiers: (i) $\beta=0$: no regularization; this corresponds to using $m(x)$ as the detection score, (ii) oracle that assumes access to labels for out-of-distribution samples in the traffic data with fixed regularization that rely on the oracle labels ($\beta=0$ for in-distribution samples and $\beta=1$ for out-of-distribution samples), (iii) $\beta=\beta(x)$: our proposed adaptive regularization with $\gamma=10$, and (iv) GAN: GAN based regularization~\cite{lee2017training}. We observe the following results:

\begin{table}[h]
\centering
\begin{small}
\scalebox{1.0}{
\begin{tabular}{c|cccc|cccc}
Dataset & \multicolumn{4}{c} {AUPR} & \multicolumn{4}{c} {Detection accuracy} \\
&Standard&Oracle&Adaptive&GAN&Standard&Oracle&Adaptive&GAN\\\hline
CIFAR-in/SVHN-out &.877 &.872 &.876 &.873 & .680&.999&.999&.750 \\
SVHN-in/CIFAR-out &.976&.968 &.970 &.974 & .780 &.998 &.997 &.950 \\
\end{tabular}}
\end{small}
\vspace{0.2in}
\caption{The results are identical to those observed in Table 1 when the traffic data was comprised exclusively of out-of-distribution data: the proposed adaptive regularization scheme again performs close to the standard classifier wrt classification performance on in-distribution data, and close to an oracle that has access to labeled out-of-distribution samples wrt out-of-distribution sample detection.}
\end{table}

From these results, it is clear that the observations from the extreme settings (traffic data is fully in-distribution or out-of-distribution) we showcased previously carry over to this mixed setting where the traffic has both in and out of distribution data - the adaptive regularization method continues to perform close to a classifier with no regularization for classification on in-distribution data, while simultaneously being extremely accurate when detecting out-of-distribution samples.

\subsection{Online training}
Finally, we consider the online training scenario where the model is trained initially only on in-distribution labeled data, and then fine-tuned in an online fashion with incoming traffic. This style of training is appropriate when the data is non-stationary, and the model needs to be updated periodically to account for the changing nature of the out-of-distribution data. In this study, we restrict our attention to the case where CIFAR-10 is the in-sample dataset and SVHN is the out-of-sample dataset.

For this experiment, we trained the classifier without any regularization ($\beta=0$) for 40 epochs, and then subsequently fine-tuned the classifier over an additional 5 epochs using the adaptive regularization scheme on a traffic dataset with an even mixture of in and out-distribution data. For the classifier trained with adaptive regularization in this online fashion, the classification AUPR dropped marginally from 0.873 to 0.865 and the detection accuracy dropped ever so slightly from 0.999 to 0.995 relative to a classifier trained with adaptive regularization on traffic data from scratch over a set of 40 epochs.

\paragraph{Analysis of $\beta(x)$:} Next, we do an analysis of the $ \beta(x)$ values as the classifier goes from epoch 40 (no-regularization) to epoch 45 (adpative regularized classifier). At the end of epoch 40, where the classifier has been trained with no regularization, the values of $\beta(x)$ have a mean and standard deviation of 0.0114 and 0.0388 for the in-distribution data, and 0.0024 and 0.0109 for the out-of-distribution data. From these figures, we can see that the $\beta(x)$ scores for the standard classifier trained with no regularization has some signal wrt detection of out-of-distribution data, but there is clear overlap in the $\beta(x)$ scores between the  in-distribution and out-of-distribution samples. This overlap is corroborated by the $68\%$ accuracy of the standard classifier in Table 1.

However, by the end of fine-tuning after epoch 45, the distribution of $\beta(x)$ changes significantly. For in-distribution data, the average value of $\beta$ is 0.52, with a standard deviation of 0.02. The corresponding figures for out-of-distribution data are 0.014 and 0.006. Now, there is clear separation between the distribution of the $\beta(x)$ values for in-distribution and out-of-distribution data. This transition from overlapping $\beta(x)$ values to values with clear separation is remarkable, but not surprising given the 99\% detection accuracy of the adaptive regularized classifier.

\paragraph{Feedback effect:} We ran an additional experiment where we froze the values of $\beta(x)$ after epoch 40, and then fine-tuned the classifier with these fixed values of $\beta(x)$ for epochs 41 to 45. In this case, our technique produced a detection accuracy of 90\%, which is still significantly better than the detection accuracy of 68\% of the standard classifier with no regularization (at the end of epoch 40), but is worse than the classifier where the values of $\beta(x)$ are allowed to change through epochs 41 to 45 (detection accuracy of 99\%). This seems to indicate that the idea of circularly using the classifiers maximum probability scores to adapt the regularization weight has a positive feedback effect that significantly improves detection accuracy. The positive direction of the feedback loop through the use of adaptive regularization is evidenced by the fact that the detection accuracy is boosted from 68\% under no regularization to 90\% when using static $\beta(x)$ weights from the no regularization classifier to fine-tune the adaptive classifier. We plan to investigate this effect further, and in particular investigate if there are failure modes where this circular feedback can harm as opposed to help performance. 



\section{Conclusion}
We proposed an adaptive regularization scheme that leverages the live traffic that hits a classifier in practice to increase the robustness to out-of-distribution samples. The key idea is to use adaptive regularization to maximize the entropy of out of distribution samples in the traffic more severely than in-distribution samples. This is achieved using the maximum probability score output by the classifier, and can be easily incorporated into the training of classifiers. Using this maximum probability score as an adaptive regularizer exhibits a circular positive feedback effect which tremendously boosts the detection accuracy. This scheme allows for us to successfully detect out of distribution samples in the traffic, without fear of harming the classifier performance on in-distribution data. In scenarios where the out-of-distribution data is non-stationary, classifiers can be updated in regular periodic intervals in an online fashion to detect the relevant out-of-distribution samples. Finally, we note that if the sample data used for regularization during training does not match the actual out-of-distribution data - for instance when the traffic data is highly non-stationary - then the traffic agnostic methods such as the GAN based regularization in \cite{lee2017training} should be preferred.

\bibliographystyle{plain}
\bibliography{ref}

\end{document}